\documentclass[runningheads]{llncs}

\usepackage{eccv}

\usepackage{eccvabbrv}

\usepackage{graphicx}
\usepackage{booktabs}

\usepackage[accsupp]{axessibility}  

\usepackage{hyperref}

\usepackage{orcidlink}

\usepackage{enumitem}
\usepackage{graphicx}
\usepackage[utf8]{inputenc} 
\usepackage[T1]{fontenc}    
\usepackage{hyperref}       
\usepackage{url}            
\usepackage{booktabs}       
\usepackage{amsfonts}       
\usepackage{nicefrac}       
\usepackage{microtype}      
\usepackage{xcolor}         
\usepackage[table]{xcolor}
\usepackage{array}
\usepackage{amssymb}
\usepackage{enumitem}
\usepackage{graphicx}
\usepackage{subcaption}
\usepackage{threeparttable}
\usepackage{pifont}
\usepackage{amsmath}
\usepackage{multirow}

\usepackage{listings}
\usepackage{makecell}
\usepackage{hyperref}
\usepackage{url}
\usepackage{fontawesome5}

\definecolor{red}{RGB}{238, 68, 51}
\definecolor{blue}{RGB}{70, 177, 225}
\definecolor{yellow}{RGB}{255, 192, 0}
\definecolor{purple}{RGB}{216, 110, 204}
\definecolor{brown}{RGB}{127, 36, 28}
\definecolor{green}{RGB}{71, 172, 20}
\definecolor{orange}{RGB}{194,153,107}

\newcolumntype{C}[1]{>{\centering\let\newline\\\arraybackslash\hspace{0pt}}m{#1}}
\newcolumntype{L}[1]{>{\raggedright\let\newline\\\arraybackslash\hspace{0pt}}m{#1}}

\begin{document}

\title{PIRA-Bench: A Transition from Reactive GUI Agents to GUI-based Proactive Intent Recommendation Agents} 

\titlerunning{PIRA-Bench}




\author{
 \textbf{Yuxiang Chai\textsuperscript{1}},
 \textbf{Shunye Tang\textsuperscript{2}},
 \textbf{Han Xiao\textsuperscript{1}},
 \textbf{Rui Liu\textsuperscript{3}},
 \textbf{Hongsheng Li\textsuperscript{1}\textsuperscript{\textdagger}}
\\
 \small
 \textsuperscript{1}MMLab @ CUHK,
 \textsuperscript{2}Nankai University,
 \textsuperscript{3}Huawei Research,
 \textsuperscript{\textdagger}Corresponding Author
 \\
 \small
 \url{https://www.pira-bench.top}
}

\institute{}

\maketitle

\begin{abstract}
  Current Graphical User Interface (GUI) agents operate primarily under a reactive paradigm: a user must provide an explicit instruction for the agent to execute a task. However, an intelligent AI assistant should be proactive, which is capable of anticipating user intentions directly from continuous visual inputs, such as mobile or desktop screenshots, and offering timely recommendations without explicit user prompting. Transitioning to this proactive paradigm presents significant challenges. Real-world screen activity is rarely linear; it consists of long-horizon trajectories fraught with noisy browsing, meaningless actions, and multithreaded task-switching. To address this gap, we introduce PIRA-Bench (Proactive Intent Recommendation Agent Benchmark), a novel benchmark for evaluating multimodal large language models (MLLMs) on continuous, weakly-supervised visual inputs. Unlike reactive datasets, PIRA-Bench features complex trajectories with multiple interleaved intents and noisy segments with various user profile contexts, challenging agents to detect actionable events while fitting to user preferences. Furthermore, we propose the PIRF baseline, a memory-aware, state-tracking framework that empowers general MLLMs to manage multiple task threads and handle misleading visual inputs. PIRA-Bench serves as an initial step toward robust and proactive GUI-based personal assistants. 
\end{abstract}

\section{Introduction}
\label{sec:intro}

The rapid evolution of Multimodal Large Language Models (MLLMs)~\cite{bai2025qwen25vltechnicalreport, bai2025qwen3vltechnicalreport, qwen35blog} has fundamentally transformed the landscape of human-computer interaction, paving the way for sophisticated Graphical User Interface (GUI) agents~\cite{hu2025osagentssurveymllmbased, shi2026trustworthyguiagentssurvey, tang2025surveymllmbasedguiagents, liu2025llmpowered}. By leveraging strong visual understanding and reasoning capabilities, these agents can navigate complex operating systems and execute tasks across mobile and desktop environments. Recent industry and academic advancements, exemplified by integrated systems like the Doubao Phone, have demonstrated that modern GUI agents are highly capable of completing a wide array of user tasks accurately when provided with natural language instructions.

Despite these impressive capabilities, the current paradigm of GUI automation remains fundamentally reactive. Existing state-of-the-art frameworks, such as UI-TARS and UI-Venus, function primarily as passive executors. They require human users to specify instructions meticulously, often necessitating a high degree of detail to ensure successful task completion. This reliance on explicit prompting places a significant cognitive burden on the user. In dynamic, real-world scenarios, a user might easily forget specific context or omit crucial details, such as the exact time, location, or name of a restaurant during a conversation, which causes reactive agents to either fail or require tedious, step-by-step clarification.

\begin{figure}[t]
    \centering
    \includegraphics[width=0.98\linewidth]{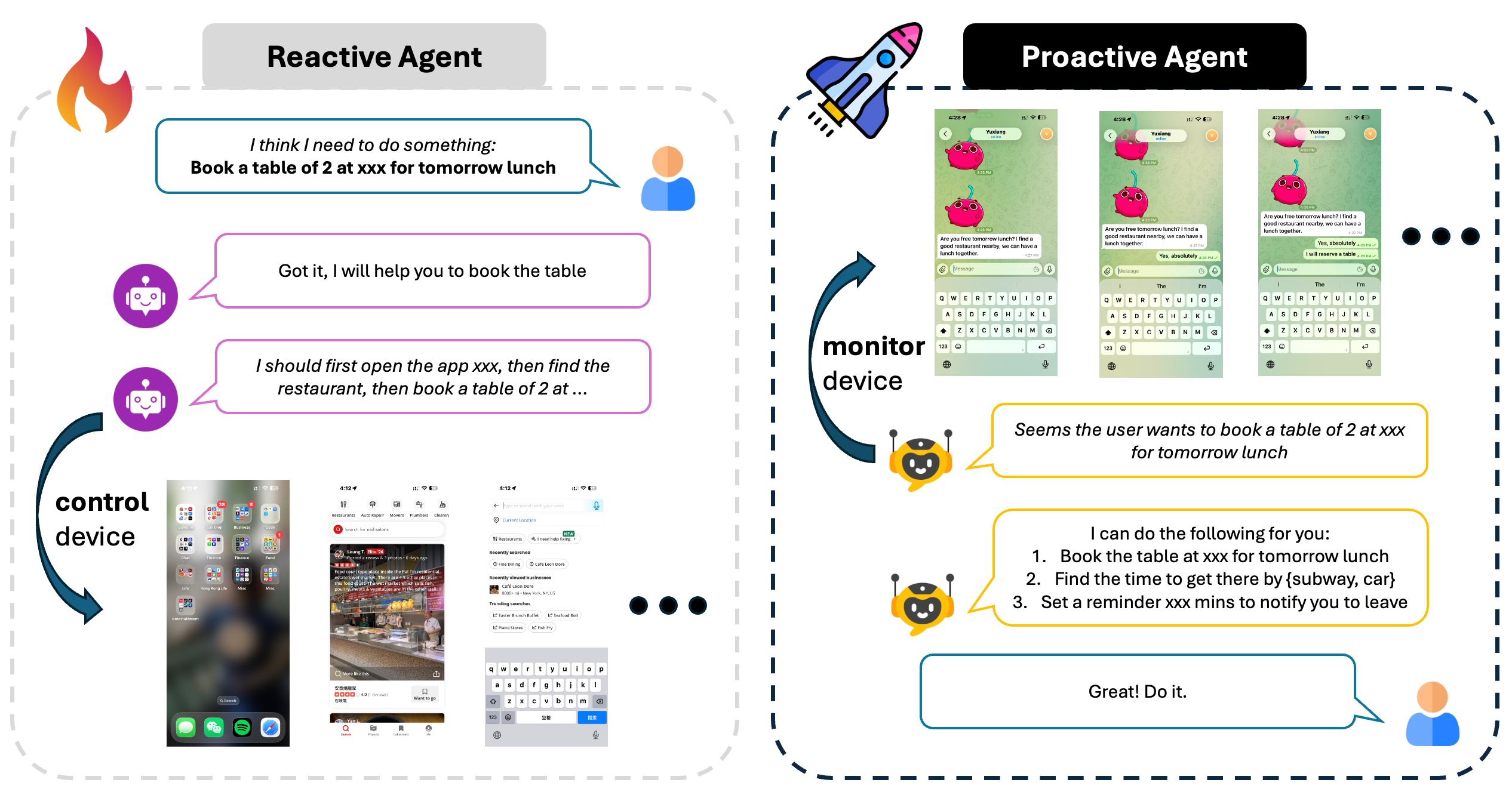}
    \caption{A comparison between the traditional reactive GUI agent paradigm and the proposed proactive intent recommendation agent. Reactive agents (left) function as passive executors, remaining idle until they receive instructions to initiate device control. In contrast, a proactive agent (right) continuously monitors the visual context to autonomously infer the user's latent goals before the user even formulates a command.}
    \label{fig:proactive}
\end{figure}

To bridge the gap between reactive automation and a "true" AI assistant, agents must evolve to anticipate user needs without requiring explicit, highly detailed prompts. We introduce the concept of the Proactive Intent Recommendation (PIR) agent, illustrated in Fig~\ref{fig:proactive}. A PIR agent continuously monitors a contextual stream of GUI screenshots (a sequence of $N$ images) from a device and autonomously predicts what the user will likely want to do next. For example, if a user is chatting with a friend about planning a meal for the upcoming weekend, a PIR agent will seamlessly analyze the visual context, extract the relevant details, and proactively recommend a set of actionable instructions, such as booking a table at the mentioned restaurant, setting a reminder, and adding a calendar event. By independently formulating the exact instructions the user intended but had not yet typed, the agent acts as a comprehensive, forward-thinking assistant.

To catalyze research and development in this promising new direction, we propose PIRA-bench, a novel benchmark designed to systematically evaluate Proactive Intent Recommendation in current models and agents. PIRA-bench comprises 100 annotated trajectories, with each trajectory containing an average of 32 sequential screenshots and paired with 3 distinct user profiles. To comprehensively assess an agent's reasoning capabilities, the benchmark evaluates performance across three scenario types: direct intent recommendation, profile-dependent prediction, and noise rejection. In direct recommendation scenarios, the agent must infer the user's future actions relying exclusively on the visual context without profile distraction. Crucially, to reflect complex human multitasking, these trajectories frequently contain multiple interleaved and crossed intents. For instance, a user might alternate between chatting with a friend about a weekend meal and studying course materials; the agent must successfully disentangle these concurrent activities to proactively recommend a composite set of actions, such as booking a restaurant and creating a calendar event, alongside summarizing study materials. The second scenario introduces deep personalization, where the agent must condition its recommendations on specific user profiles. For example, given a similar housing search trajectory, an agent should recommend buying premium real estate for a wealthy user, while recommending affordable rentals for a student. Finally, to rigorously simulate real-world conditions where users are often idle or distracted, we insert noises to those trajectories and also include trajectories composed entirely of pure noise. This requires agents to not only extract hidden intents from cluttered streams but also strictly demonstrate operational restraint by preventing false predictions when no actionable intent exists.

To establish a strong foundation for this challenging task, we further propose the Proactive Intent Recommendation Framework (PIRF). PIRF serves as a baseline designed to empower general MLLMs to iteratively process long sequences of visual context. The framework features a dedicated memory module that dynamically records and tracks ongoing multitasking states and specific user profile contexts. To effectively mitigate hallucinations, which is a common failure mode when general models process noisy trajectories, PIRF incorporates a built-in reflection mechanism. By continuously evaluating the memorized tasks and executing an auto-deletion process for outdated or completed intents, PIRF maintains a clean, accurate representation of the user's true goals.

To summarize, our main contributions are three-fold:
\begin{itemize}[label=$\bullet$]
    \item We introduce the Proactive Intent Recommendation (PIR) task, shifting the focus of GUI agents from purely reactive, instruction-following executors to forward-thinking assistants capable of anticipating user needs from continuous visual streams.
    \item We construct PIRA-bench, a comprehensive dataset containing 100 real-world, multi-step trajectories with 3 user profiles each (averaging 33 screenshots). The benchmark is uniquely designed with interleaved multi-tasking scenarios, user profile context, and intentional noise to rigorously evaluate an agent's ability to disentangle concurrent activities and filter out distractions.
    \item We propose PIRF, a novel architecture that equips general MLLMs with iterative processing capabilities, a dynamic memory module, and a reflection-based auto-deletion mechanism. This framework effectively tracks complex, crossed intents and significantly reduces hallucinations when processing lengthy, noisy visual trajectories.
\end{itemize}

\section{Related Work}

\subsection{GUI Agent}

The rapid advancement of Multimodal Large Language Models (MLLMs)~\cite{bai2025qwen3vltechnicalreport, bai2025qwen25vltechnicalreport, kimiteam2026kimik25visualagentic, qwen35} has catalyzed significant progress in the development of Graphical User Interface (GUI) agents~\cite{hu2025osagentssurveymllmbased, tang2025surveymllmbasedguiagents, shi2026trustworthyguiagentssurvey, liu2025llmpowered}. These systems are designed to perceive visual screen states and execute complex action sequences across mobile and desktop operating systems. Recent pioneering works have heavily optimized the instruction-following and visual-grounding capabilities of these agents. For instance, UI-TARS~\cite{qin2025uitarspioneeringautomatedgui, wang2025uitars2technicalreportadvancing} and Mobile-Agent-V3/3.5~\cite{ye2025mobileagentv3fundamentalagentsgui, xu2026mobileagentv35multiplatformfundamentalgui} have demonstrated robust, end-to-end navigation and task completion on smartphone interfaces, effectively mapping natural language commands to precise screen interactions. Similarly, models like UI-Venus~\cite{venusteam2026uivenus15technicalreport, gu2025uivenustechnicalreportbuilding}, UI-Genie~\cite{xiao2025uigenie} and MAI-UI~\cite{zhou2025maiuitechnicalreportrealworld} have introduced enhanced capability for better UI element localization and action planning, significantly improving the success rates of complex, multi-step tasks. Furthermore, the integration of advanced reinforcement learning and post training, as seen in models like InfiGUI-R1~\cite{liu2025infiguir1advancingmultimodalgui}, UI-R1~\cite{lu2025uir1enhancingefficientaction} and GUI-R1~\cite{luo2025guir1generalistr1style}, allows agents to have stronger reasoning capabilities.

\subsection{General AI Assistant}

Beyond the specialized domain of GUI navigation, the broader trajectory of artificial intelligence research is shifting from passive tools toward autonomous, proactive personal agents. This paradigm shift is heavily exemplified by recent open-source initiatives such as OpenClaw\footnote{https://openclaw.ai/}. Operating as a self-hosted, continuous automation engine, OpenClaw bridges the gap between digital environments and general task execution by integrating Large Language Models directly with local operating systems and cross-platform messaging APIs. Unlike traditional chatbots restricted to isolated, stateless sessions, these agents maintain persistent memory and operate continuously in the background. They demonstrate the viability of autonomous execution, managing calendars, summarizing communications, and running scheduled background tasks based on historical context without requiring immediate human prompting. In the academic pursuit of this proactive ideal, recent literature has begun exploring intent awareness on digital devices~\cite{cohen-etal-2025-small, yang2025fcmirmobilescreenawareness}. For instance, the FC-MIR framework~\cite{yang2025fcmirmobilescreenawareness} leverages screen context to detect user goals and recommend relevant actions. However, FC-MIR fundamentally focuses on identifying the intent of a task the user is currently executing or actively interacting with to accelerate in-progress workflows. Our proposed Proactive Intent Recommendation (PIR) task diverges significantly by focusing on the inference of future, latent goals. For example, a PIR agent must autonomously propose booking a restaurant based on the context of an ongoing messaging chat, rather than merely assisting within a restaurant application the user has already explicitly opened.

\section{Proactive Intent Recommendation}

Graphical User Interface (GUI) agents have emerged as powerful tools for digital automation, significantly reducing the manual effort required to navigate complex software ecosystems. By accurately mapping natural language instructions to executable system actions, these agents excel at completing well-defined tasks, from web browsing to application management. However, their utility is fundamentally bottlenecked by their reactive nature. Users are required to explicitly formulate detailed instructions, a process that imposes a cognitive burden and interrupts the natural flow of work. In real-world scenarios, users frequently multitask, switch between applications, or simply forget specific context, such as the exact time of an appointment or the name of a venue mentioned in a passing conversation. To transcend the limitations of passive execution, there is a critical need for Proactive Intent Recommendation (PIR) agents. Rather than waiting for explicit commands, PIR agents continuously observe the user's visual context, anticipate their latent goals amidst noise and distractions, and autonomously propose timely, actionable tasks. This shift from instruction-following to intent-anticipation represents the crucial next step in evolving GUI agents into comprehensive, intelligent assistants.

\subsection{Reactive GUI Agent Paradigm}

Standard GUI agents operate as instruction-following executors within a sequential decision-making framework, typically modeled as a Partially Observable Markov Decision Process (POMDP). Formally, given a specific, user-provided natural language instruction $I$, an initial visual state $s_0$, and a predefined action space $\mathcal{A}$, the agent's objective is to generate a sequence of actions to fulfill $I$. At any time step $t$, the agent receives a visual observation $s_t$ (a GUI screenshot) and samples an action $a_t \in \mathcal{A}$ from its policy $\pi$:
$$a_t \sim \pi(a \mid s_t, I, \mathcal{H}_{<t})$$
where $\mathcal{H}_{<t} = (s_0, a_0, s_1, a_1, \dots, s_{t-1}, a_{t-1})$ represents the historical trajectory of states and actions. In this paradigm, the instruction $I$ is a strictly necessary prerequisite; without explicit human prompting, the policy cannot determine the goal state, rendering the agent idle.

\subsection{Proactive Intent Recommendation (PIR) Paradigm}

In contrast, a comprehensive AI assistant must anticipate user needs before an explicit instruction $I$ is given. We define the Proactive Intent Recommendation task as the continuous analysis of a passive observation stream to predict the user's latent future goals, as illustrated in Fig~\ref{fig:proactive}. Let $\mathcal{T} = (s_1, s_2, \dots, s_N)$ denote a trajectory of $N$ sequential GUI screenshots passively captured from the user's device. Let $\mathcal{P}$ denote the user profile, encapsulating personalized preferences and socio-economic status. The objective of a PIR agent is to learn a mapping function $f_\theta$ that predicts a set of future, actionable intents $\mathcal{I}^* = \{i_1, i_2, \dots, i_K\}$ that the user is highly likely to execute or desire next. We formulate this mapping as finding the intent set that maximizes the conditional probability $P_\theta$:$$\mathcal{I}^* = f_\theta(\mathcal{T}, \mathcal{P}) = \arg\max_{\mathcal{I}} P_\theta(\mathcal{I} \mid \mathcal{T}, \mathcal{P})$$
where each intent $i_k \in \mathcal{I}^*$ is a natural language instruction or structured command (e.g., "book a table for two at Grnd Restaurant at 7 PM").


\subsection{Real-World Complexity: Interleaved Intents and Noise}

Real-world trajectories are rarely monolithic. Due to user multitasking and distractions, the trajectory $\mathcal{T}$ is often a disjointed union of sub-trajectories corresponding to multiple concurrent tasks, alongside irrelevant noise. We can formulate the observed trajectory as:
$$\mathcal{T} = \mathcal{T}_{task_1} \cup \mathcal{T}_{task_2} \dots \cup \mathcal{T}_{task_M} \cup \mathcal{T}_{noise}$$
where $\mathcal{T}_{task_m}$ represents the subset of non-contiguous frames relevant to a specific latent intent $i_m$, and $\mathcal{T}_{noise}$ represents frames resulting from app-switching, random browsing, or idle scrolling. Therefore, a successful PIR agent must not only maximize $P(\mathcal{I} \mid \mathcal{T}, \mathcal{P})$ but also perform temporal credit assignment and disentanglement to successfully map interleaved state subsets ($\mathcal{T}_{task_1}, \dots, \mathcal{T}_{task_M}$) to their corresponding distinct intents, while aggressively driving the probability of generating intents from $\mathcal{T}_{noise}$ to zero.

\section{PIRA-Bench}

While the rapid advancement of Multimodal Large Language Models (MLLMs) has spurred the creation of numerous environments and datasets for GUI automation, the research community currently lacks a standardized evaluation framework dedicated to proactive assistance. Existing benchmarks and mobile GUI datasets predominantly focus on reactive, instruction-following capabilities. In these traditional setups, the evaluation metric is strictly tied to how accurately an agent executes a pre-defined, explicit natural language command. Consequently, current benchmarks or environments are structurally unsuited to assess an agent's ability to infer latent, future goals from a passive, continuous stream of visual context.

To accurately measure and drive progress in this paradigm shift, it is imperative to establish a rigorous evaluation standard. We need a framework that tests not just what a model can do when told, but how well it can anticipate user needs, handle interleaved multi-tasking, and filter out real-world noise to propose actionable tasks. To address this critical gap, we introduce \textbf{PIRA-Bench}, the first benchmark explicitly designed to evaluate Proactive Intent Recommendation Agents. By shifting the evaluation criteria from instruction-execution accuracy to intent-prediction relevance, PIRA-Bench provides the foundational infrastructure necessary to systematically assess, compare, and improve the proactive capabilities of current and future MLLMs.

\subsection{Dataset Composition}

To systematically assess the proactive capabilities of MLLMs in a realistic environment, PIRA-Bench is constructed with a total of 100 meticulously curated GUI trajectories. These trajectories represent sequential, passive observation streams captured from diverse real-world device usage, encompassing both mobile phone and desktop environments. Rather than isolating capabilities into rigid subsets, the dataset presents a unified, mixed distribution of challenges that mirrors the unpredictability of actual user behavior.

To ensure robust evaluation across different dimensions of proactivity, the 100 trajectories are designed with the following universal and variable features:

\begin{itemize}[label=$\bullet$]
    \item \textbf{Universal Noise Injection and Profiling}: To simulate the clutter of realistic usage, \textit{every} trajectory in the dataset is injected with noise frames, such as irrelevant app switching, idle screens, or random browsing at leisure, which requires agents to continuously filter out distractions. Furthermore, to enable scalable personalization testing, each of the 100 trajectories is paired with three distinct user profiles, encapsulating varying socio-economic statuses, preferences, and characteristics.
    \item \textbf{Spectrum of Intent Dependencies}: Within the dataset, the trajectories cover a spectrum of dependency on user context:
        \begin{enumerate}
            \item \textit{Direct Recommendation Cases}: In these instances, the visual context provides sufficient information to infer the user's future goals (e.g., "Schedule a meeting based on the chat content"). The latent intents are broadly applicable and can be predicted directly from the screenshots, testing the agent's ability to disentangle interleaved tasks.
            \item \textit{Profile-Dependent Cases}: These trajectories present scenarios where the visual context alone is ambiguous or insufficient. Here, the agent must cross-reference the visual cues with the assigned user profile to generate the correct intent (e.g., distinguishing whether a user intends to "buy a luxury apartment" or "rent a budget studio" based on financial constraints), thereby evaluating deep personalization capabilities.
        \end{enumerate}
    \item \textbf{Negative Rejection Samples (Pure Noise)}: To rigorously test operational safety and hallucination resistance, a portion of the trajectories are designed as negative samples. Despite being paired with user profiles and containing valid GUI interactions, these sequences consist entirely of aimless actions or fragmented browsing with \textit{no} actionable latent goal. For these cases, the ground-truth expectation is for the agent to correctly identify the absence of intent and strictly propose no action.
\end{itemize}

By integrating these diverse scenarios, ranging from context-rich tasks to profile-dependent decisions and pure noise, into a single unified dataset, PIRA-Bench provides a holistic testbed for evaluating the predictive accuracy, personalization capability, and operational restraint of proactive agents.

\subsection{Evaluation}

To ensure a robust and standardized assessment of proactive agents, PIRA-Bench employs a rigorous ground-truth annotation process coupled with an automated, scalable evaluation pipeline using state-of-the-art models.

\subsubsection{Ground Truth Annotation}

Establishing objective ground truth (GT) for latent intents is inherently challenging due to the subjective nature of human goals and the specific constraints of user profiles. To mitigate individual bias, we employ a consensus-based human annotation strategy. For each of the 100 trajectories, three independent humans are provided with the visual sequence and the specific user profile. They separately annotate the explicit, actionable intents they infer. The final GT intent set, denoted as $\mathcal{I}_{\text{GT}}$, is constructed by aggregating these annotations and retaining only the intents that reach a majority agreement (i.e., identified by at least two out of the three annotators). For trajectories designed as pure noise or where the profile constraints negate any actionable task, the consensus GT correctly maps to an empty set ($\mathcal{I}_{\text{GT}} = \emptyset$).

\subsubsection{LLM-as-a-Judge Evaluation}

Evaluating the exact lexical match of predicted intents against the GT is sub-optimal. Therefore, we adopt an LLM-as-a-judge paradigm. Gemini-3-flash is utilized to semantically compare the agent's predicted intent set, $\mathcal{I}_{\text{pred}}$, against $\mathcal{I}_{\text{GT}}$. Crucially, the judge is also provided with the user profile context to supplementarily verify that the predicted intents satisfy the specific socio-economic and preference constraints of the user.

\subsubsection{Metrics Formulation}

To comprehensively capture an agent's ability to accurately anticipate profile-aligned intents while resisting hallucinations, we define three core metrics based on the nature of the ground truth:

\begin{itemize}[label=$\bullet$]
    \item \textbf{Average Intent F1 Score ($\text{F1}_{\text{avg}}$):} This metric assesses the model's accuracy on all trajectories where actionable intents exist (i.e., $\mathcal{I}_{\text{GT}} \neq \emptyset$). We compute the F1 score independently for each such trajectory $j$ by comparing the semantic alignment of $\mathcal{I}_{\text{pred}}^{(j)}$ and $\mathcal{I}_{\text{GT}}^{(j)}$. The final score is the macro-average across all positive samples:
    $$\text{F1}_{\text{avg}} = \frac{1}{|T_{\text{pos}}|} \sum_{j \in T_{\text{pos}}} \text{F1}^{(j)}$$
    where $T_{\text{pos}}$ represents the set of trajectories containing valid latent intents.

    \item \textbf{Normalized False Positive Score ($\text{FPS}_{\text{norm}}$):} To rigorously assess robustness, we evaluate performance on all trajectories where no actionable intent exists. We first calculate the False Positive Score ($\text{FPS}$) as the average number of hallucinated intents per negative sample. To map this unbounded count to a normalized metric $\text{FPS}_{\text{norm}} \in (0, 1]$, we apply a logarithmic dampening function:
    $$\text{FPS}_{\text{norm}} = \frac{1}{1 + \ln(1 + \text{FPS})}$$
    This formulation mimics the diminishing returns of error perception, which strictly penalizes the initial hallucinations while preventing the metric from vanishing too rapidly for models with moderate noise levels. A perfect model ($\text{FPS} = 0$) achieves a score of $1.0$.

    \item \textbf{Final Score ($\mathcal{S}_{\text{final}}$):} To provide a unified assessment of an agent's overall reliability, we compute the Final Score as the product of the proactive capability and the hallucination penalty:
    $$\mathcal{S}_{\text{final}} = \text{F1}_{\text{avg}} \cdot \text{FPS}_{\text{norm}}$$
    By formulating the Final Score multiplicatively, $\text{FPS}_{\text{norm}}$ acts as a strict reliability scaling factor. This ensures that a top-performing agent must simultaneously master complex, profile-aware intent prediction and maintain operational restraint in noisy environments.
\end{itemize}

\section{Proactive Intent Recommendation Framework (PIRF)}
\label{sec:pirf}

\begin{figure}[t]
    \centering
    \includegraphics[width=0.98\linewidth]{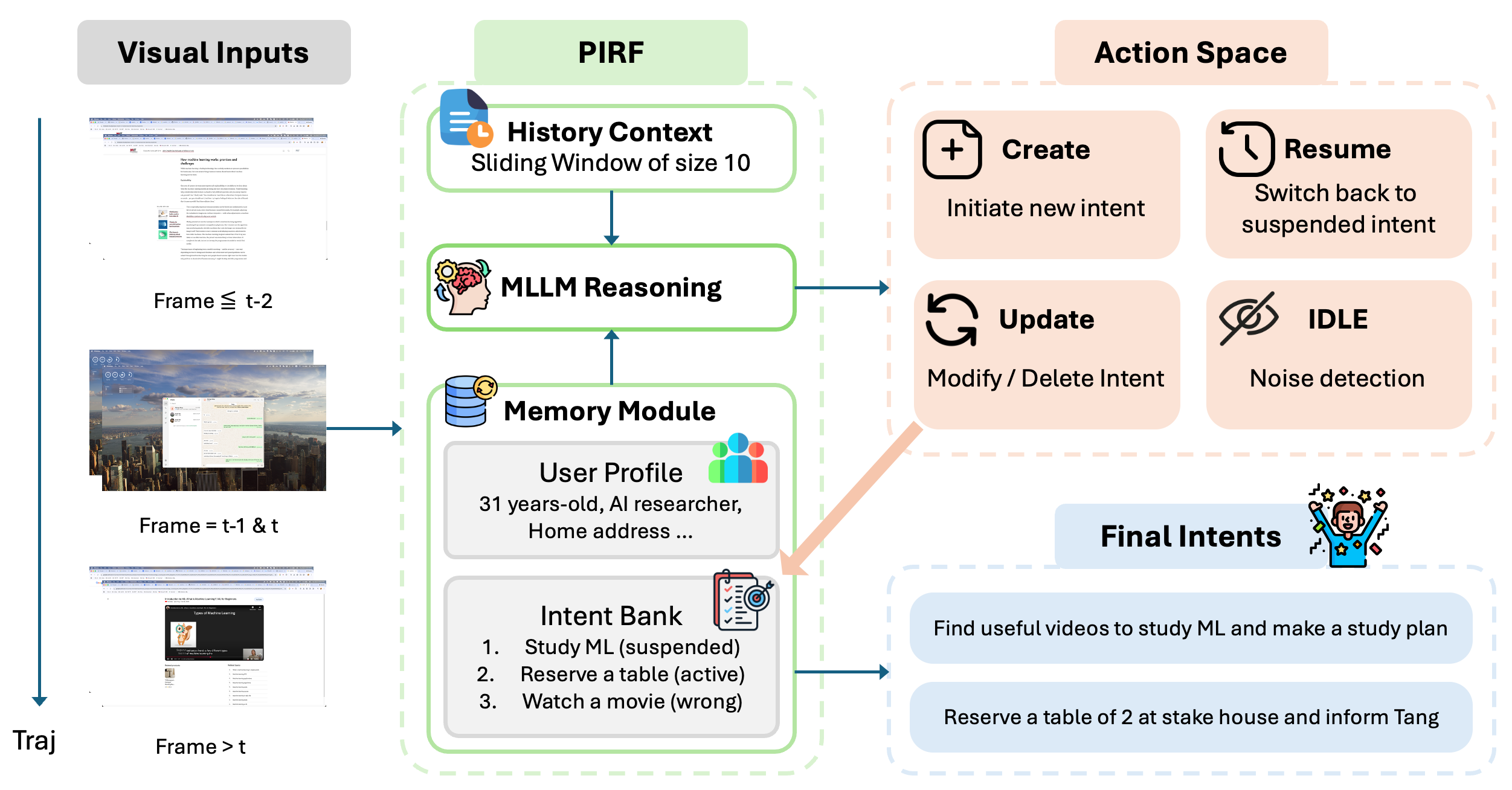}
    \caption{The overall architecture of the Proactive Intent Recommendation Framework (PIRF). The system processes continuous visual inputs sequentially by integrating a sliding window of historical context with a dynamic Memory Module that maintains both static user profiles and a continuously updated intent bank. Driven by a Multimodal Large Language Model, the framework evaluates the current state against a structured action space (i.e., Create, Resume, Update, or IDLE) to autonomously refine its memory and generate a final set of actionable, proactive intents.}
    \label{fig:pirf}
\end{figure}

To establish a robust baseline for the PIRA-Bench dataset, we propose the Proactive Intent Recommendation Framework (PIRF). Standard Multimodal Large Language Models, while highly capable of single-turn visual understanding, often struggle with long-horizon context retention and are highly susceptible to hallucinations when presented with noisy, continuous visual streams (see in Section~\ref{sec:exp}). PIRF addresses these limitations by wrapping a general MLLM in a structured, state-tracking cognitive architecture designed specifically for continuous intent disentanglement and reflection, as illustrated in Fig.~\ref{fig:pirf}.

\subsection{Architecture and Dynamic Memory Module}

At its core, PIRF treats the intent recommendation task as a continuous state-updating process. Instead of processing the entire trajectory $\mathcal{T}$ at once, which is computationally expensive and inapplicable to the real-world scenario, PIRF processes the visual stream sequentially. To handle the complex, interleaved multi-tasking scenarios in PIRA-Bench and the real world, PIRF employs a multifaceted dynamic Memory Module. This module anchors the user profile $\mathcal{P}$, capturing essential personalization constraints, socio-economic context, and individual preferences. Concurrently, it maintains a dynamic list of active "threads," where each thread represents a distinct, suspended user intent. At any time step $t$, the framework injects this combined structured memory state, which pairs the static user profile with the ongoing tasks (e.g., Intent 1: Study machine learning, Intent 2: Reserve a table), directly into the MLLM's context. To provide immediate temporal context without maintaining a huge context window, PIRF also utilizes a sliding conversational window, retaining only the $K$ most recent frames and reasoning steps (where $K=10$ in our baseline implementation).

\subsection{Intent Action Space and State Transitions}

At each frame, the framework prompts the underlying MLLM to analyze the current visual observation against the Suspended Intents memory and output a structured state transition. The action space is defined as follows:
\begin{itemize}[label=$\bullet$]
    \item \texttt{CREATE}: Triggered when the visual context indicates the initiation of a novel task. PIRF instantiates a new thread ID and generates an abstract, proactive intent description.
    \item \texttt{RESUME}: Triggered when the user switches back to a previously suspended task. The framework updates its active pointer to the corresponding thread ID, effectively disentangling interleaved multi-tasking.
    \item \texttt{UPDATE}: Triggered when the current screen represents a continuation or progression of the currently active intent, allowing the framework to refine the intent description as more context is revealed.
    \item \texttt{IDLE}: A critical action for hallucination mitigation. When the framework detects that the current screen is meaningless noise (e.g., idle scrolling, application homepages, or random browsing), it outputs IDLE. This explicit rejection class prevents the model from over-triggering and ensures safety on the pure noise subsets of PIRA-Bench.
\end{itemize}

\subsection{Reflection and Auto-Deletion Mechanism}

A significant challenge in processing long GUI trajectories is memory bloat: as users abandon tasks due to context changing, stale intents remain in the context window, confusing the model and degrading the average intent F1 score. To combat this, PIRF incorporates a Reflection and Auto-Deletion Mechanism. Independent of the primary state transitions (CREATE, RESUME, etc.), the framework enforces a continuous reflection protocol at every time step. The model must explicitly evaluate if the visual evidence suggests any intent in the memory bank has been either modified or abandoned. If so, it issues a \texttt{delete\_intent\_id} parameter. PIRF immediately purges these obsolete threads from the active memory. This auto-deletion ensures that the cognitive load remains low and the predicted intent pool strictly reflects the user's current latent goals, maximizing the final reliability score $\mathcal{S}_{\text{final}}$.

\section{Experiments}
\label{sec:exp}

\subsection{Settings}
We evaluate the performance of state-of-the-art Multimodal Large Language Models (MLLMs) on PIRA-Bench using two distinct experimental settings (i.e., a naive baseline and our proposed Proactive Intent Recommendation Framework (PIRF)) alongside a human performance reference:

\begin{itemize}[label=$\bullet$]
    \item \textbf{Naive MLLM Baseline}: To assess the inherent proactive capabilities of current models without specialized architecture, we employ a standard "sliding context" approach. In this setting, the MLLM is fed a sequence of $N=10$ resized screenshots at each turn. To serve as a naive baseline, the model is instructed via the system prompt to "remember the context and observe user behavior" for the initial frames and is strictly forbidden from outputting predictions until the final turn of the window. This baseline tests the model's raw ability to retain visual context and identify intents without external memory modules or structured state tracking.
    \item \textbf{PIRF (Ours)}: We evaluate the models within the Proactive Intent Recommendation Framework (PIRF) described in Section~\ref{sec:pirf}. In this setup, the models utilize the dynamic Memory Module to track suspended intents and the Reflection Mechanism to auto-delete completed tasks. The context window is similarly maintained at 10 frames, but the model is queried at every step to update its memory state (CREATE, RESUME, UPDATE) or remain IDLE, rather than just at the end of the sequence.
    \item \textbf{Human Performance}: To establish an empirical upper bound and validate the solvability of the benchmark, we conducted a human performance study. Human evaluators were presented with the same sequential visual streams and user profiles as the models. They were asked to identify actionable intents and determine when no action was required (noise). This baseline serves as a reference for the ideal balance between intent discovery (Recall) and hallucination resistance (F1 and FPS).
\end{itemize}

For each MLLM setting, we conduct our evaluation on four leading MLLMs to represent a diverse range of capabilities: Gemini-3.1-Pro, GPT-5.2, Qwen3.5-Plus~\cite{qwen35blog}, Seed-1.8~\cite{seed2025seed1}. The quantitative results for the Naive Baseline, PIRF, and Human Performance are presented in Table \ref{tab:main_results}. 

\begin{table*}[t]
\centering
\caption{Main results on PIRA-Bench. We report the average Precision, average Recall, and average F1 ($\text{F1}_{\text{avg}}$) for intent detection, alongside the Normalized False Positive Score ($\text{FPS}_{\text{norm}}$) for noise robustness. The Final Score ($\mathcal{S}_{\text{final}}$) is the product of $\text{F1}_{\text{avg}}$ and $\text{FPS}_{\text{norm}}$. All values are reported as percentages (\%).}
\label{tab:main_results}
\resizebox{\textwidth}{!}{
\renewcommand{\arraystretch}{1.25}
\setlength{\tabcolsep}{4pt}
    \begin{tabular}{C{1.7cm}| L{2.5cm} | C{2cm}|C{1.5cm}|C{1.5cm}|C{2cm}|C{1.5cm}}
        \toprule
        \textbf{Method} & \textbf{Model} & \textbf{Precision} & \textbf{Recall} & \textbf{$\text{F1}_{\text{avg}}$} & \textbf{$\text{FPS}_{\text{norm}}$} & \textbf{$\mathcal{S}_{\text{final}}$} \\
        \midrule
        \multirow{4}{*}{\textbf{Naive}} & Gemini-3.1-Pro & 48.57 & 69.97 & 45.08 & 49.61 & 22.36 \\
            & GPT-5.2        & 31.95 & 83.37 & 40.75 & 31.31 & 12.76 \\
            & Qwen3.5-Plus   & 40.47 & 69.36 & 46.58 & 45.46 & 21.18 \\
            & Seed-1.8       & 43.53 & 68.74 & 47.68 & 48.85 & 23.29 \\
        \midrule
        \multirow{4}{*}{\textbf{PIRF}} & Gemini-3.1-Pro & \textbf{53.05} & 78.97 & \textbf{56.58} & 45.39 & 25.68 \\
            & GPT-5.2        & 50.52 & \textbf{84.54} & 54.68 & 43.90 & 24.00 \\
            & Qwen3.5-Plus   & 42.07 & 70.21 & 49.87 & 47.60 & 23.73 \\
            & Seed-1.8       & 51.82 & 72.67 & 55.71 & \textbf{50.36} & \textbf{28.05} \\
        \midrule
          \textbf{Human} & - & 98.76 & 89.67 & 93.89 & 96.23 & 90.35 \\
        \bottomrule
    \end{tabular}
}
\end{table*}

The performance of GPT-5.2 in the Naive setting serves as a stark illustration of the "over-proactivity" trap. While it achieves a remarkable recall of 83.37\%, which is the highest among all naive baselines, this sensitivity proves deceptive. The accompanying precision is critically low (31.95\%), and its noise robustness score ($\text{FPS}_{\text{norm}}$) is the worst in the table at 31.31\%. This pattern characterizes a "trigger-happy" model: it correctly identifies most true user intents simply by aggressively predicting intents from many noise frames. In a real-world proactive assistant, this behavior is catastrophic; the utility of correct suggestions is drowned out by a flood of hallucinations during idle moments. Consequently, despite having high recall, its final score ($\mathcal{S}_{\text{final}}$) is penalized heavily (12.76), accurately reflecting that a hyper-active assistant is functionally unusable.

From the table, a notable observation is the consistent superiority of the PIRF framework over the Naive baseline across all evaluated models. By integrating a structured dynamic memory module and a reflection mechanism, PIRF significantly improves the final score ($\mathcal{S}_{\text{final}}$) for every model. For GPT-5.2, PIRF acts as a cognitive filter. While recall remains high (increasing slightly to 84.54\% due to better context tracking), the reflection mechanism drives a dramatic improvement in precision (+18.57 points to 50.52\%) and noise robustness (+12.59 points to 43.90\%). This indicates that the framework successfully disentangles "eagerness" from "accuracy," allowing the model to retain its reasoning power while learning to remain silent during noise.

A comparison between Gemini-3.1-Pro and Seed-1.8 highlights a fundamental trade-off in proactive agent design: Gemini-3.1-Pro (PIRF) achieves the highest $\text{F1}_{\text{avg}}$ (56.58\%) and precision (53.05\%). It can formulate the most accurate intent descriptions when tasks are present. Seed-1.8 (PIRF), however, achieves the highest final score (28.05). This victory is driven by its superior $\text{FPS}_{\text{norm}}$ (50.36\%). This result suggests that for the PIRA-Bench metric, operational restraint is weighted as heavily as capability. Seed-1.8 acts as a "conservative" agent: it misses some intents (lower Recall of 72.67\%), but it rarely hallucinates during noise. In the context of an always-on background assistant, this conservative behavior yields a higher overall reliability score than a more capable but "noisier" model. The data suggests that future research must focus not just on boosting recall, but on teaching frontier models when not to act.

The human performance baseline reveals a substantial gap between current MLLMs and human capability, with the human $\mathcal{S}_{\text{final}}$ of 90.35 far surpassing the best model performance of 28.05 (Seed-1.8). Crucially, this disparity is driven not merely by the ability to identify intents, where GPT-5.2's recall of 84.54\% approaches the human benchmark of 89.67\%, but by the profound difference in operational restraint. Humans demonstrate near-perfect precision (98.76\%) and noise robustness (96.23\%), indicating an almost flawless ability to distinguish true latent goals from background noise. In contrast, even the most capable models struggle significantly with false positives, underscoring that the primary challenge for future proactive agents is not just increasing sensitivity, but mastering the discernment to remain silent when no action is required. However, this superior discernment comes at a significant cost, as human testers required approximately 15 to 20 times the inference time of the models to process the same trajectories.

\subsection{Ablation Study}

To quantify the "distraction cost" of real-world visual clutter, we conducted an ablation study analyzing the impact of noise injection on agent performance. In this experiment, we compared the performance of the PIRF framework across two dataset variations on the positive samples which have intents:
\begin{itemize}[label=$\bullet$]
    \item \textbf{Clean Trajectories}: The original sequences with all noise frames (idle scrolling, etc.) manually removed, leaving only the intent-relevant frames.
    \item \textbf{Noised Trajectories (Standard)}: The standard PIRA-Bench evaluation subset, containing the full mixture of relevant actions and injected noise.
\end{itemize}

\begin{table*}[t]
\centering
\caption{Ablation study on the impact of noise injection using the PIRF framework. We compare performance on \textbf{Clean} trajectories (noise frames removed) versus \textbf{Noised} trajectories (standard PIRA-Bench). All values are reported as percentages (\%).}
\label{tab:ablation}
\resizebox{0.85\textwidth}{!}{
\renewcommand{\arraystretch}{1.25}
\setlength{\tabcolsep}{4pt}
    \begin{tabular}{C{2cm}| L{2.5cm} | C{2cm}|C{2cm}|C{2cm}}
        \toprule
        \textbf{Traj} & \textbf{Model} & \textbf{Precision} & \textbf{Recall} & \textbf{$\text{F1}_{\text{avg}}$} \\
        \midrule
        \multirow{4}{*}{\textbf{Noised}} & Gemini-3.1-Pro & 53.05 & 78.97 & 56.58   \\
            & GPT-5.2        & 50.52 & 84.54 & 54.68  \\
            & Qwen3.5-Plus   & 42.07 & 70.21 & 49.87  \\
            & Seed-1.8       & 51.82 & 72.67 & 55.71  \\
        \midrule
        \multirow{4}{*}{\textbf{Clean}} & Gemini-3.1-Pro & 85.28 & 74.44 & 76.42  \\
            & GPT-5.2        & 92.23 & 83.57 & 84.46  \\
            & Qwen3.5-Plus   & 81.78 & 72.62 & 74.02  \\
            & Seed-1.8       & 83.34 & 71.86 & 74.39  \\
        \bottomrule
    \end{tabular}
}
\end{table*}

The ablation study results in Table \ref{tab:ablation} isolate the "distraction cost" of real-world visual environments. By comparing performance on Clean trajectories (where only intent-relevant frames are retained) versus Noised trajectories (with noise frames injected), we observe a distinct failure mode in current MLLMs: high susceptibility to visual distraction. The most significant finding is the drastic collapse in precision across all models when noise is introduced. GPT-5.2, which achieves a near-perfect Precision of 92.23\% on clean trajectories, plummets to 50.52\% on Noised trajectories, which is a degradation of over 40 percentage points. Similarly, Gemini-3.1-Pro drops from 85.28\% to 53.05\%. This indicates that while current MLLMs are highly capable of interpreting user behavior in idealized, noise-free environments (Clean $\text{F1}_{\text{avg}} > 74\%$), they lack the robustness to function reliably in the wild. The introduction of irrelevant frames (e.g., idle scrolling or random browsing) confuses the models, causing them to misinterpret noise as meaningful trigger signals and hallucinate intents that do not exist.

Counter-intuitively, the Recall for the top-performing models (Gemini-3.1-Pro and GPT-5.2) actually increases slightly in the Noised setting (e.g., GPT-5.2 improves from 83.57\% to 84.54\%). This phenomenon suggests that noise triggers an "over-proactive" behavior. Confronted with a cluttered visual stream, these powerful models tend to lower their detection thresholds and generate more predictions to ensure they don't miss potential tasks. While this strategy successfully captures the true intents (maintaining or boosting Recall), it comes at the devastating cost of Precision described above. The models effectively become "trigger-happy," failing to exercise the restraint necessary for a helpful assistant.

\section{Conclusion}

In this work, we introduce PIRA-bench, a novel benchmark designed to evaluate GUI-based proactive assistants. Through 100 diverse, real-world trajectories paired with distinct user profiles and realistic visual noise, we provide the first systematic assessment of an agent's ability to autonomously discover latent intents, disentangle interleaved tasks, and personalize recommendations. To address these complexities, we propose the Proactive Intent Recommendation Framework (PIRF), a baseline architecture integrating dynamic memory with reflection. Our extensive evaluation reveals that while frontier models demonstrate high recall, they suffer from significant "over-proactivity" and hallucinate intents when confronted with visual noise. However, the consistent gains achieved by PIRF demonstrate that structured state tracking and self-reflection are viable pathways to mitigate these failures, establishing a rigorous standard for future research into agents that are not only smarter but also more discerning about when to act.

%
%
\bibliographystyle{splncs04}
\bibliography{main}
\end{document}